\documentclass[letterpaper]{article} % DO NOT CHANGE THIS
\usepackage{aaai24}  % DO NOT CHANGE THIS
\usepackage{times}  % DO NOT CHANGE THIS
\usepackage{helvet}  % DO NOT CHANGE THIS
\usepackage{courier}  % DO NOT CHANGE THIS
\usepackage[hyphens]{url}  % DO NOT CHANGE THIS
\usepackage{graphicx} % DO NOT CHANGE THIS
\usepackage{natbib}  % DO NOT CHANGE THIS AND DO NOT ADD ANY OPTIONS TO IT
\usepackage{caption} % DO NOT CHANGE THIS AND DO NOT ADD ANY OPTIONS TO IT
\usepackage{algorithm}
\usepackage{algpseudocode} 
\usepackage{newfloat}
\usepackage{listings}
\DeclareCaptionStyle{ruled}{labelfont=normalfont,labelsep=colon,strut=off} % DO NOT CHANGE THIS
\floatstyle{ruled}
\newfloat{listing}{tb}{lst}{}
\floatname{listing}{Listing}
\usepackage{tabularx}% NEW ADDED
\usepackage{multirow}% NEW ADDED
\usepackage{amsmath}% NEW ADDED
\usepackage{tabularray}% NEW ADDED
\usepackage{array}% NEW ADDED
\usepackage{booktabs}% NEW ADDED
\usepackage{subfigure}% NEW ADDED

\title{DocMSU: A Comprehensive Benchmark for Document-level Multimodal  \\Sarcasm Understanding}
\author{
    Hang Du\textsuperscript{\rm 1},
    Guoshun Nan\textsuperscript{\rm 1}\thanks{Guoshun Nan is the corresponding author.},
    Sicheng Zhang\textsuperscript{\rm 1},
    Binzhu Xie\textsuperscript{\rm 1},
    Junrui Xu\textsuperscript{\rm 1},
    Hehe Fan\textsuperscript{\rm 2},
    Qimei Cui\textsuperscript{\rm 1},
    Xiaofeng Tao\textsuperscript{\rm 1},
    Xudong Jiang\textsuperscript{\rm 3}
    }
\affiliations{
  \textsuperscript{\rm 1}Beijing University of Posts and Telecommunications, China\\
  \textsuperscript{\rm 2}Zhejiang University\\
  \textsuperscript{\rm 3}Nanyang Technological University, Singapore\\
 \{7597892, nanguo2021, zhangsicheng, xbz\_nicous, bbxst0371\}@bupt.edu.cn, crane.h.fan@gmail.com, \{cuiqimei, taoxf\}@bupt.edu.cn, EXDJiang@ntu.edu.sg
}

\begin{document}

\maketitle

\begin{abstract}
Multimodal Sarcasm Understanding (MSU) has a wide range of applications in the news field such as public opinion analysis and forgery detection. 
However, existing MSU benchmarks and approaches usually focus on sentence-level MSU. 
In document-level news, sarcasm clues are sparse or small and are often concealed in long text. 
Moreover, compared to sentence-level comments like tweets, which mainly focus on only a few trends or hot topics (e.g., sports events), content in the news is considerably diverse.  
Models created for sentence-level MSU may fail to capture sarcasm clues in document-level news. 
To fill this gap, we present a comprehensive benchmark for Document-level Multimodal Sarcasm Understanding (DocMSU). 
Our dataset contains 102,588 pieces of news with text-image pairs, covering 9 diverse topics such as  health, business, etc.
The proposed large-scale and diverse DocMSU significantly facilitates the research of document-level MSU in real-world scenarios. 
To take on the new challenges posed by DocMSU, we introduce a fine-grained sarcasm comprehension method to properly align the pixel-level image features with word-level textual features in documents. 
Experiments demonstrate the effectiveness of our method, showing that it can serve as a baseline approach to the challenging DocMSU.
Our code and dataset are available at https://github.com/Dulpy/DocMSU.
\end{abstract}

\section{Introduction}
\label{sec:intro}
Sarcasm is a form of verbal irony that often uses positive words to convey a negative message, such as frustration, anger, contempt and even ridicule \cite{WILSON20061722}. In real-world cases,  a piece of sarcastic news often lacks explicit linguistic markers, and thus requires additional cues to reveal the true intentions.  The accompanying visual information provides helpful cues to better perceive ironic discrepancies. Multimodal sarcasm \cite{wang-etal-2022-multimodal,KaiShu2017FakeND} is omnipresent in social media posts, forum discussions, and product reviews, and hence the multimodal sarcasm understanding is of great significance for a wide range of applications in the news field such as sentiment analysis\cite{Mao_Shen_Yu_Cai_2021}, fake news detection\cite{Ying2022BootstrappingMR,Qi_Bu_Cao_Ji_Shui_Xiao_Wang_Chua_2023}, and public opinion analysis.     
\begin{figure}
    \centering
    \includegraphics[width=76mm]{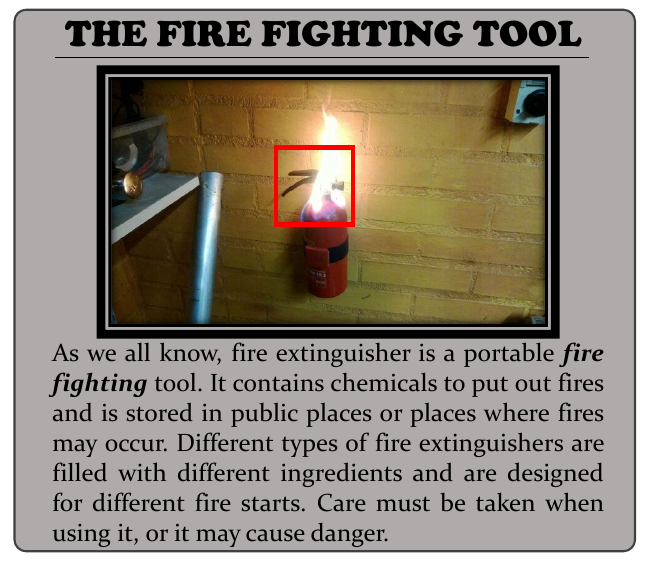}
    \caption{Example of a piece of sarcastic news. It shows that a fire extinguisher, a tool used to extinguish fires, has caught fire. This sarcasm can remind us of the quality of the fire extinguisher.}
    \label{fig:introexample}
    \vspace{-7mm}
\end{figure}

Figure \ref{fig:introexample} illustrates a piece of multimedia sarcastic news. 
To understand this sarcastic news, a model must capture textual cues from multiple sentences, including \textit{fire fighting} and \textit{cause danger}. The accompanying visual cue, \textit{fire on a fire extinguisher} in the figure below, plays an important role in this sarcasm.
The figurative and creative nature of such multimodal sarcasm poses a great challenge to the effective perception of the true intention under the guise of overt positive surface involved in a whole document and an image. This requires the design and development of document-level multimodal sarcasm understanding (MSU) methods that specifically take the characteristics of such an ironic expression into consideration.  

% Previous works have demonstrated the great importance of leveraging large, high-quality and challenging benchmarks to develop and evaluate the state-of-the-art deep learning methods for various natural language processing (NLP) tasks \cite{8269806,10.1007/978-3-030-58577-8_8}. Along this line, existing sarcasm benchmarks \cite{castro-etal-2019-towards,wang-etal-2022-multimodal} have shown their promise. However, towards document-level MSU in the real-world news field, they have some limitations, 
Prior research has underscored the critical significance of utilizing extensive, high-quality, and challenging benchmarks for the development and evaluation of state-of-the-art deep learning methods across various natural language processing (NLP) tasks\cite{8269806,10.1007/978-3-030-58577-8_8}. In this context, existing sarcasm benchmarks \cite{castro-etal-2019-towards,wang-etal-2022-multimodal} have demonstrated considerable promise. However, when addressing document-level multimodal sarcasm understanding in the real-world news domain, they exhibit certain limitations, including (1) Limited length of text. In real-world scenarios, a piece of news may include more than $70$ words across multiple sentences \cite{KaiShu2017FakeND}, concealing ironic discrepancies beyond sentence boundaries. However, samples in existing multimodal sarcasm datasets \cite{cai-etal-2019-multi,castro-etal-2019-towards,wang-etal-2022-multimodal} only include about $20$ words within a single utterance on average, which greatly simplifies the challenges of MSU in real-world cases. (2) Limited quality of annotations. Existing large satirical datasets \cite{riloff-etal-2013-sarcasm,ptacek-etal-2014-sarcasm,barbieri-etal-2014-modelling} are mostly generated by bootstrapping algorithm or remote supervision with noisy labels. These annotations can be disruptive to systems that harness such data for downstream applications due to the subtle nature of sarcasm. Furthermore, these datasets only contain text modality. (3) Very limited number of samples. As sarcasm lacks explicit linguistic or visual markers, a model requires a large volume of samples to learn the rules or ways that reveal the true underlying intentions. A large-scale dataset benefits the generalization capability of an MSU model that alleviates the over-fitting issue during the training procedure.

The aforementioned limitations in existing datasets highlight the need for a comprehensive, challenging, and higher-quality document-level multimodal sarcasm dataset to enhance irony understanding in the domain of news. Towards that, we developed DocMSU, a comprehensive benchmark that contains high-quality annotations of $102,588$ pieces of news with text-image pairs, covering $9$ hot topics such as science, business, and sports. We collect these samples from social websites, including ``New York Times'' and ``UN News'', each involving $63$ tokens across $5$ sentences on average. To alleviate the ambiguity of sarcasm, we manually annotated these documents and images in $3$ rounds with $15$ workers, ensuring the annotation quality with confidence scores. Each pair of text-image involves a binary label for sarcasm detection, $2.7$ textual spans and visual bounding boxes on average for sarcasm localization.  

The proposed DocMSU facilitates the research of multimodal sarcasm perception for real-world applications. It also introduces two new challenges: (1) capturing the nuanced sarcastic clues in two modalities, where the clues are concealed within very few words in a document or a tiny area in an image;
 (2) aligning the visual and linguistic features for irony understanding, where the incongruity nature of sarcasm requires cross-modal interactions. To fill this gap, we propose a novel sarcasm comprehension method that aims to fuse the pixel-level image features with the word-level textual features of a whole document in a fine-grained manner. Experimental results show the effectiveness of our method. We will release our dataset and the code. The main contributions of our work can be summarised as follows:

%are concealed beyond   sentences, and  are able to reflect the sarcasm, the sarcasm may hide in detail of the image as well;

%Based on our dataset, we conduct 3 tasks to further promote the understanding of sarcasm. Due to the poor performance of existing multimodal methods, we proposed a novel multimodal method, which aims to align different modalities for tasks with fine-grained features. Because current evaluation matrix cannot evaluate the model's performance accurately. We newly designed three evaluation matrix to meet the requirements. 

\begin{itemize}
	\item We curate DocMSU, a new benchmark  for document-level multimodal sarcasm understanding in the real-world news field. Compared with existing ones, our dataset is more comprehensive and more challenging with much higher quality annotations.
% To the best of our knowledge, this is the first large, challenging and high-quality dataset for irony understanding in the domain.
	\item We come up with a novel document-level MSU method for sarcasm detection and localization, mitigating the issues in sarcastic cues detection across sentences and across modalities under inconsistent context.     
% to localize sarcasm objects in both texts and images based on our dataset, and introduce three evaluation matrix to accurately estimate the model's performance.
	\item We conduct extensive experiments on our DocMSU. Results show that the created benchmark enables us to develop and evaluate various deep learning methods for the task of MSU closer to the real-world application.
\end{itemize}

\section{Related Work}
% \HH{1. Vision-Language Tasks/Applications (datasets + methods) 2. MSU (datasets + methods): may need to highlight the difference between MSU and existing vision-language tasks.}
%Sarcasm zunderstanding has gained increasing interests in recent years, but the datasets of this field are limited. 

%In this section, we first investigate existing sarcasm datasets, and then discuss the methods proposed for sarcasm understanding.

\noindent\textbf{Datasets}: Existing sarcasm datasets are mainly collected from Twitter and Reddit and can be roughly categorized into text-based ones  \cite{riloff-etal-2013-sarcasm,ptacek-etal-2014-sarcasm,barbieri-etal-2014-modelling,khodak-etal-2018-large,oprea-magdy-2020-isarcasm}, and the multimodal ones \cite{cai-etal-2019-multi,castro-etal-2019-towards,wang-etal-2022-multimodal}. The text-based datasets suffer from noisy labels caused by remote supervision. The most related to our work is MSTI  \cite{wang-etal-2022-multimodal}. However, the texts in MSTI only contain $20$ tokens on average, which may not well reflect challenges in the news field. Compared to the existing sarcasm datasets, our DocMSU provides more samples, much longer texts and higher quality annotations towards sarcasm understanding in practice of the real-world news field. Detailed comparisons are available in Table \ref{tab:comparsion}. 

\noindent
\textbf{Methods}: 
Early studies of sarcasm understanding were based on statistical patterns \cite{riloff-etal-2013-sarcasm,joshi-etal-2015-harnessing} and deep learning techniques such as word embeddings and LSTM/CNN  \cite{joshi-etal-2015-harnessing,zhang-etal-2016-tweet}. Recent MSTI leverages pre-trained BERT and ResNet to extract the cross-modal features\cite{wang-etal-2022-multimodal}. Some powerful methods such as CLIP \cite{radford2021learning} and VILT\cite{hu2019local} rely on contrastive learning and Transformer to learn multimodal representations. Different from the above methods, our model aims to comprehend the fine-grained nuanced sarcastic clues in two modalities, where the clues reside within very few words in a document or a very tiny area in an image.
%beyond sentence boundary and learn fine-grained features for modality alignment.    
%\par Here we combine the characteristics of DocMSU to propose a novel method to settle the new challenges from our dataset, we align different modalities between each word in the text and each pixel in the image to get fine-grained features. The results of experiments prove that our model boosts the performance of object localization in both texts and images.
\begin{table*}[]

\resizebox{\linewidth}{!}{
\scalebox{0.98}{
\begin{tabular}{ccccccc}
\hline
\textbf{Datasets} & \textbf{Volume}  & \textbf{Level} & \textbf{Source}& \textbf{Input} & \textbf{Labeled Obj.}& \textbf{Annotator} \\ \hline
Riloff \cite{riloff-etal-2013-sarcasm} & 175,000 & Sentence & Tweets &  Text & - & Manual \\
iSarcasm \cite{oprea-magdy-2020-isarcasm}  & 4,484 & Sentence & Social media  & Text &Text & Manual \\
Cai \cite{cai-etal-2019-multi}  & 24,635 & Sentence & Tweets  & Text, Image & Text& Auto \\
MSTI \cite{wang-etal-2022-multimodal}  & 5,015  & Sentence & Tweets &Text, Image & Text, Image & Manual \\
MUStARD \cite{castro-etal-2019-towards}  & 690  & Sentence & TV shows & Text, Audio, Video & - & Manual \\ \hline
\textbf{DocMSU (ours) } & \textbf{102,588} & \textbf{Document} & \textbf{News}  &Text, Image & \textbf{Text, Image} & Semi-Manual \\ \hline
\end{tabular}}
}
\caption{The comparisons between our DocMSU dataset and previous ones.}
\label{tab:comparsion}
\vspace{-3mm}
\end{table*}

\section{The DocMSU Dataset}
We present DocMSU, a new benchmark that contains high-quality annotations of $102,588$ pieces of news with text-image pairs in $9$ hot topics. 
%To enable the exploration of multimodal sarcasm detection in news field, we introduce a new dataset (DocMSU), consisting of images and texts manually annotated for their sarcasm property.
\subsection{Data Collection}
\par We crawl data from some famous news websites such as ``New York Times'', ``UN News'', ``The Onion'' and ``NewsThumb'', etc. To avoid regulation issues, we discard news that includes sensitive topics such as pornography and violence. Finally, we collect  more than $70,000$ pieces of news that consist of titles, abstracts, images, and news bodies, where each sample is generated by combing a news title, the abstract, and the image. Each sample involves $63$ tokens across $5$ sentences on average. We categorize these data into $9$ groups such as ``science'', ``health'', and ``business'', and each group involves $10$ visual object types such as ``building'', ``animal'' and ``art''. We use an open-source tool doccano \cite{doccano} for textual and visual annotations and $15$ volunteers participated in the work. We mix up different categories of samples before the annotation, allowing each annotator to randomly access news. 

%%Note that, although some pieces of news in our initial pool are obtained through the organizations that provide satirical news specifically, the news is still free of sarcasm.

\subsection{Annotation Process}
\label{sec:ann-process}
%Our ultimate goal is to produce a dataset that can be used for sarcasm target recognition in both images and texts. Through the analysis of sarcasm and tasks, there are two problems in the annotation process. 
During the annotation procedure, we give a binary tag for each document-image pair to indicate whether it is an ironic message. For a piece of sarcasm news, we further mark the sarcastic clues, including the textual span in the document and the bounding box in the image. However, we face two challenges in such an annotation procedure.
\begin{itemize}
    \item \textbf{Lacking explicit linguistic and visual markers in a sample.} An annotator may not be able to accurately understand the sarcasm in some news titles, images and the corresponding abstracts, as they may require some proper background knowledge for the annotator to understand the sarcasm.
    \item \textbf{Annotation variances caused by the subjective nature of perceiving sarcasm.} As irony is always conveyed in a subtle way both in a document or an image, the perception of sarcastic clues varies from different annotators. 
\end{itemize}

%the abstract of news 
%As the title of news cannot meet the requirements of document-level, the abstract of news cannot fully express the meaning of news, and the article of news often contains redundant information such as advertisement, which is not relevant to the news itself. (2) Due to the quality of the news collected varies greatly, the sarcasm semantic in news has a certain degree of ambiguity, and we want our annotations of news could be acknowledged by the majority of the people. Here, we come up with two solutions to resolve these issues.
%To tackle the above two issues, we introduce two methods as follows.

\begin{figure}[h!]
    \vspace{-2mm}
    \centering
    \subfigure[]{
    \includegraphics[width=72mm]{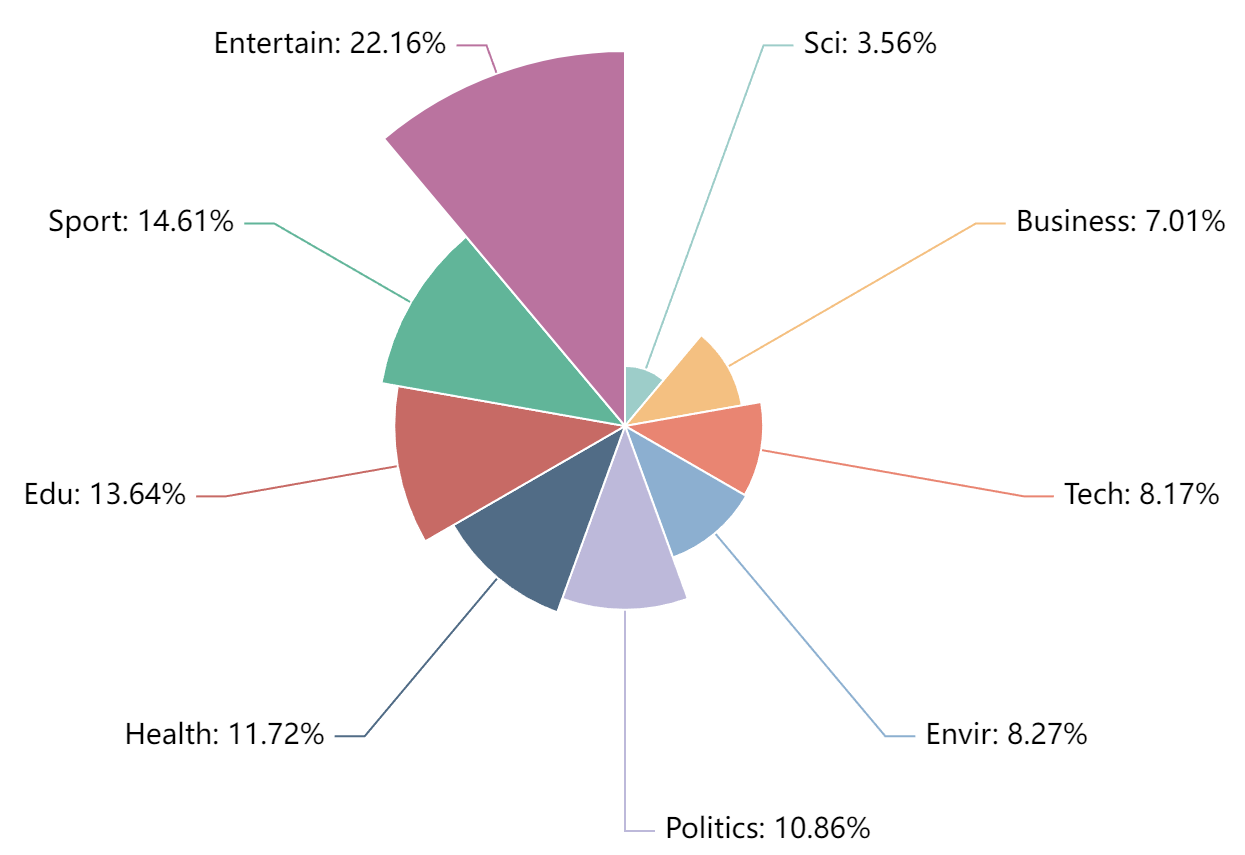}
    \label{fig:sub-a}
    }%
    %\vspace{1em}
    
    \subfigure[]{
    \includegraphics[width=72mm]{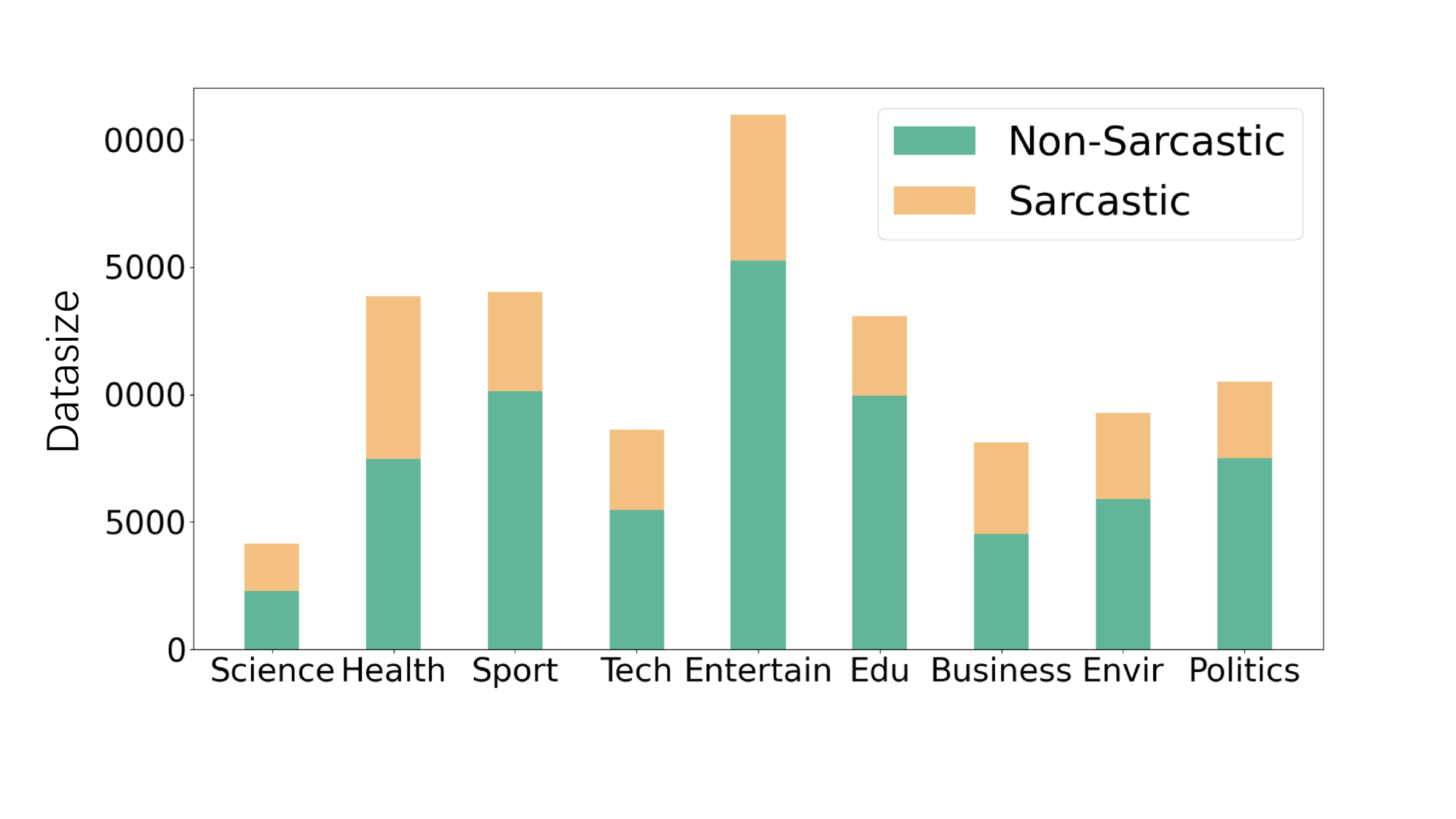}
    \label{fig:sub-b}
    }%
    
    \subfigure[]{
    \includegraphics[width=72mm]{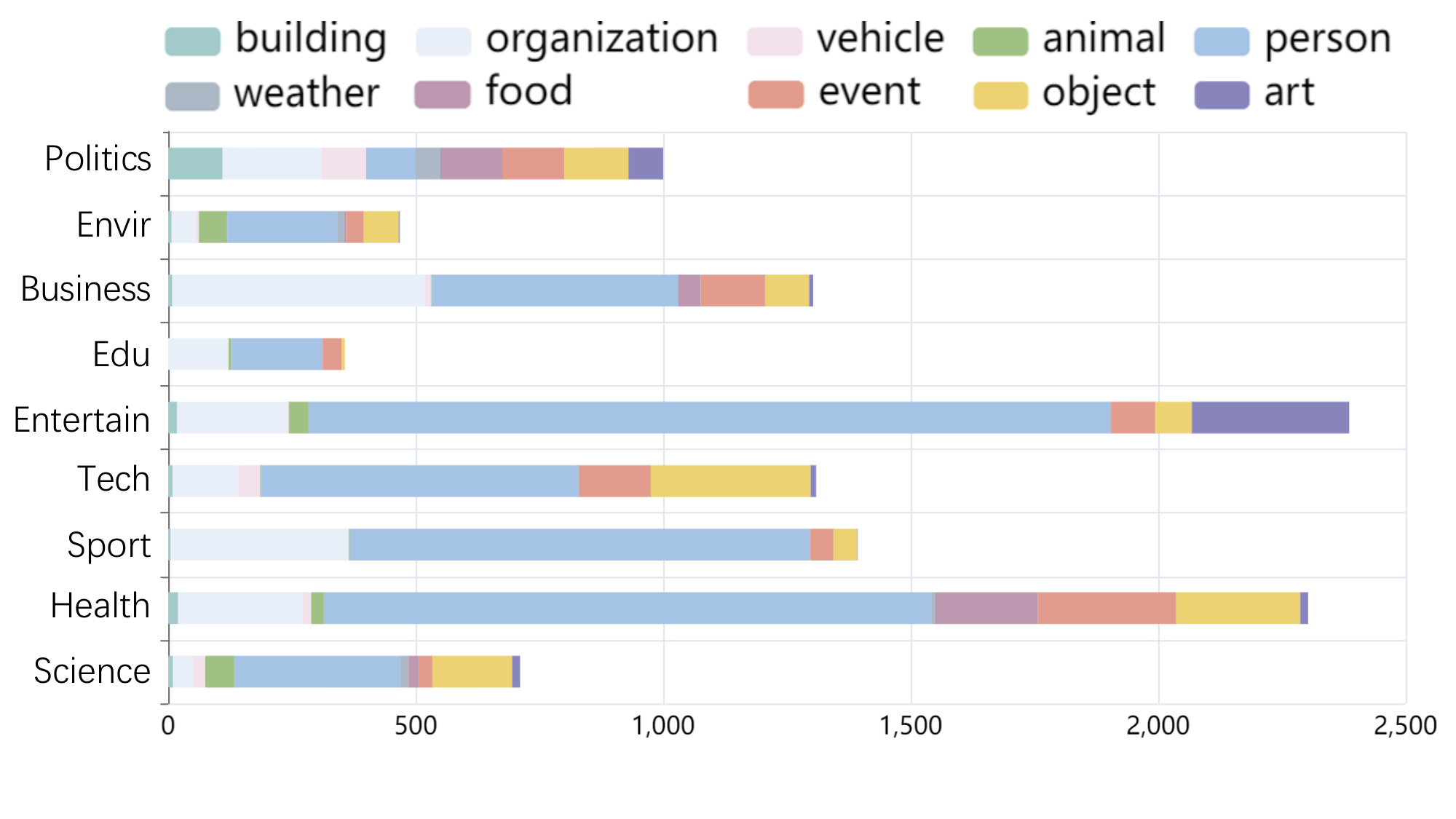}
    \label{fig:sub-c}
    }%
    
    \caption{Statistics of our DocMSU: (a) percentage of each topic in the overall dataset. (b) distribution of sarcastic samples and non-sarcasm ones in each news topic. (c) distribution of visual object type in each topic.}
    \label{fig:statics}
    \vspace{-8mm}
\end{figure}

% \begin{figure*}[t]
% 	\centering
% 	\subfigure[]{
%      \includegraphics[width=0.2\linewidth]{imgs/type_distributionv3.png}
%         \label{fig:sub-a}
% 	}
% 	\subfigure[]{
%   \includegraphics[width=0.3\linewidth]{imgs/bar.png}
% 	\label{fig:sub-b}
%         }
% 	\subfigure[]{
%      \includegraphics[width=0.3\linewidth]{imgs/obj_distribute_v2.png}
%         \label{fig:sub-c}
% 	}%
% 	\centering
% 	\caption{Statistics of our DocMSU: (a) percentage of each topic in the overall dataset. (b) distribution of sarcastic samples and non-sarcasm ones in each news topic. (c) distribution of visual object type in each topic.}
% 	\vspace{-4mm}
% 	\label{fig:my_label3}
% \end{figure*}

For the first issue, we ask the annotator to refer to the news body to better understand the context. By doing so, the annotator is able to give a more accurate binary label, as well as sarcastic clues including the textual spans in the document and bounding box in the image. Regarding the second issue, we have $3$ annotators for each sarcastic sample with a scoring mechanism. We use Intersection-over-Union (IoU) to quantify the similarity between two annotations. A similarity score between two annotations is defined as the sum of textual IoU (TIoU) and visual IoU \cite{Yu2016UnitBoxAA}. TIoU is defined as follows: $S$ refers to the text labeled by the annotator, $r$ is the index of annotator, $i$ and $j$ indicate the positions of the beginning and the end of the sarcastic span respectively. 
\par{\footnotesize\begin{align}
 TIOU = \frac{min(S_{r-1}[j],S_{r}[j])-max(S_{r-1}[i],S_{r}[i])}{max(S_{r-1}[j],S_{r}[j])-min(S_{r-1}[i],S_{r}[i])}\end{align}}For each annotation, we obtain two similarity scores with the other two annotations. The sum of them is defined as the confidence score of this annotation. The annotation with the highest confidence score is selected in our DocMSU. Due to the subtle nature of sarcasm, there are some samples whose ironic clues can be hardly distinguished. For these samples, we observe that all the three confidence scores are much smaller than those of other samples, and these samples take up about $5$ percent of the data. Hence, we ask the annotator who achieves the overall highest confidence score among the $15$ volunteers in the whole annotation procedure to further label these ``challenging'' instances. We also use GPT-3.5 to augment the text data and discard instances that may include sensitive information.

\subsection{Dataset Analysis}
Figure \ref{fig:statics} details the statistics of our DocMSU.  Figure \ref{fig:sub-a} shows the percentage of the $9$ topics, ``Science'', ``Health'', ``Sport'', ``Technology'', ``Entertainment'', ``Education'', ``Business'', ``Environment'', and ``Politics'', where the ``Environment'' topic is most popular and takes the largest portion $22.16\%$. Figure \ref{fig:sub-b} illustrates the distribution of sarcastic samples and non-sarcasm ones in each topic. Totally, our benchmark contains $34,130$ sarcastic samples and $68,458$ non-sarcastic ones. Figure \ref{fig:sub-c} shows the distribution of visual object type in each topic, where multiple types of visual objects enrich the feature for sarcasm understanding. We have $10$ object types in our DocMSU. A sample contains $2.7$ labeling targets on average, which are sarcastic clues, including textual spans in a document and bounding boxes in an image\footnote{We provide more details in Appendix: DocMSU Annotation Pipeline, including annotation user interface, data samples, etc. and appendices are available in the preprint version.}.

\begin{figure}[h]
    \centering
    \includegraphics[width=72mm]{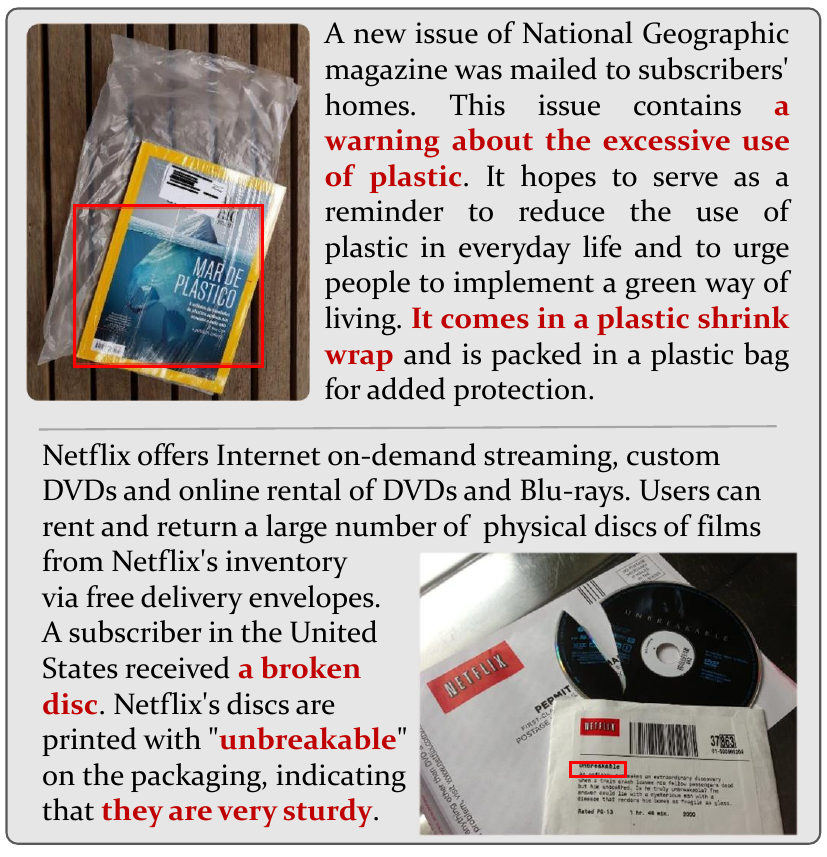}
    \caption{Two samples selected from our benchmark.}
    \label{fig:example}
\end{figure}
\begin{figure*}[ht]
    \centering
    \includegraphics[scale=0.30]{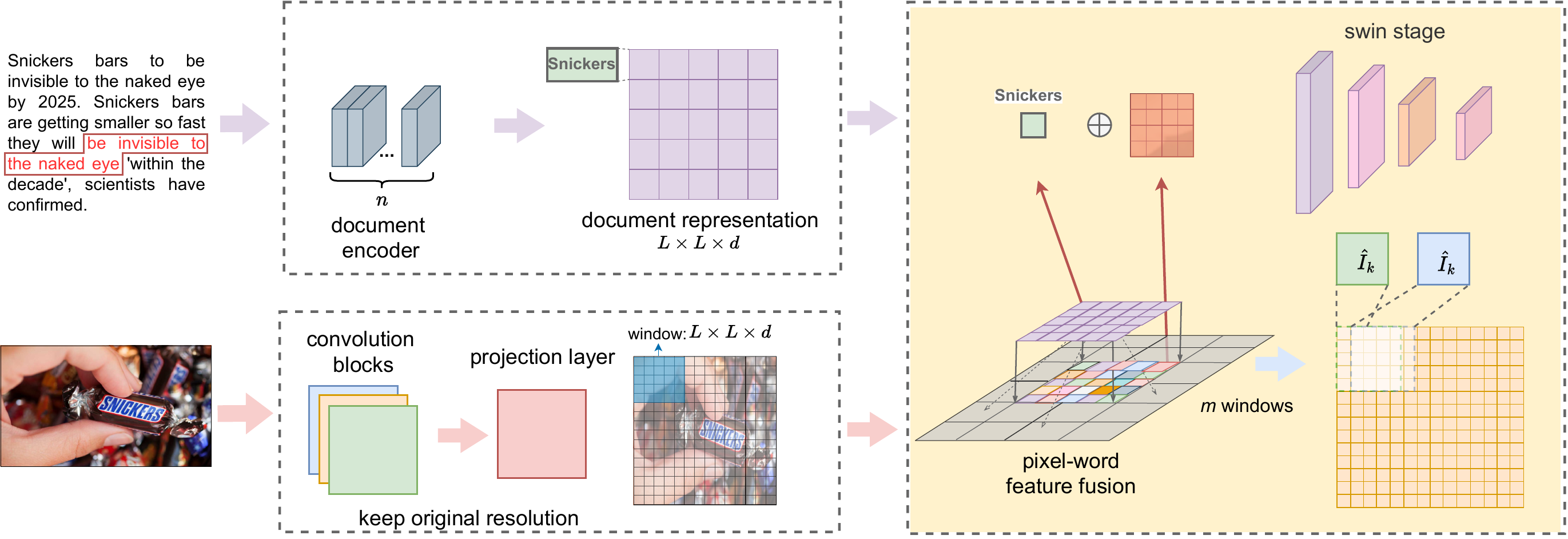}
    %\vspace{-2mm}
    \caption{Overview of the proposed model. We use the pre-trained BERT to generate contextualized token-level representations of the document and then form a document matrix of size $L \times L$ with a padding mechanism. We rely on a simplified Resnet to output image representations and a projection layer to spilt the representations of an image window into ${L \times L}$ patches. We add patches of each image window to the document matrix to fuse the two modalities. The fused representations are fed to Swin-Transformer to patch attentions with a sliding window.}
    \label{fig:model}
    \vspace{-7mm}
\end{figure*}
%\vspace{-2mm}
\section{Proposed Method}
%In this section, we first discuss the new challenges of our benchmark and then present a novel method for two MSU tasks. 
%($d \times t \times t$)%

\subsection{Motivation}
\label{sec:motivation}
Figure \ref{fig:example} shows two examples selected from our DocMSU. 
The first example highlights the irony in National Geographic's practices, as they extensively employ plastic packaging despite advocating against the overuse of plastic. The subsequent example exposes the contradictions within Netflix's service delivery. Despite boasting about their discs being ``unbreakable" and implying exceptional durability, customers received damaged discs, contradicting the advertised durability.

A model may face two new challenges for sarcasm understanding the above two examples. (1) Capturing the nuanced sarcastic clues that are concealed within very few words (e.g., ``\textit{a broken disc}'') in a document or in a very tiny area (e.g., the tag ``\textit{unbreakable}'') of the image.  (2) Aligning the visual and text features for the accurate irony understanding (e.g., ``\textit{unbreakable}'' and the broken disc). Existing approaches such as recent MSTI \cite{wang-etal-2022-multimodal}, CLIP, and VILT have limitations in tackling these two challenges as they focus more on learning the overall information of the whole text and image representations. This motivates us to develop a new method to capture the fine-grained linguistic and visual sarcastic clues and align the two different types of clues for a better MSU.

%(1) Due to the single image or text cannot thoroughly express the sarcasm, the interactions between images and documents take an important part in sarcasm understanding. (2) The textual document may include several sentences, but only a few words are able to flip the polarity of the whole document, where sarcasm exists with a high probability. 

%Image could be regarded as a document as well, each pixel is a single word in image, and only a few ``words'' in image may reflect the sarcasm semantic. 

%As shown in Figure \ref{fig:example}, ``but Trump's safe'' is the textual sarcastic cue, so as the ticket with ``made in china'' in the picture. To solve the challenges above, we need to localize the interactions of details between images and documents through multimodal fine-grained features. 

\subsection{Overview}

Figure \ref{fig:model} illustrates the architecture of our model, which mainly consists of three components, including a document encoder, an image encoder, and a fusion module. To capture the underlying subtle clues concealed within very few words in a document and a very tiny area in an image, our model generates two matrices for pixels-level image representations and token-level document representations. For the cross-modal interactions, we fuse the representations in two matrices with a sliding window for multimodal alignment. We will explore the specifics of this design in the following sections.
\subsection{Document Encoder}
%Given a sequence of words in document $s = \{w_i\}_{i=1}^{n}$, $n$ is the length of the document $s$. 

We denote a document as $s = \{w_i\}_{i=1}^{n}$, where $w_i$ indicates the $i$-th token and $n$ is the total number. We use BERT \cite{devlin-etal-2019-bert} to output contextualized token-level representations \begin{math}
\mu \in R^{n \times d} \end{math},
where
\begin{gather}
\mu = [\nu_1,\nu_2,...,\nu_n] = \mathrm{BERT}(s)
\end{gather}
We use a fully connected layer $f_c$ to transform word representations. Then, we convert the document representation into a square shape 
\begin{math}
\varpi \in R^{L \times L \times d},      
\end{math}
%As the number of words differs from each document, we apply a fully connection layer () to change the length of vectors' sequence to $L^{2}$ and reformat these vectors to document matrix .
{\begin{align}
\varpi(i,j,:) = \big(f_c(\theta)\big)_{L \times (i-1)+j}\end{align}}
where $1 \leq i,j \leq L $. 
We add paddings when $n < L \times L$. 
This square document representation is used for the fine-grained alignment of the pixel-level visual representations. 

%\mM(i,j,:) =  \sum_{0 \leq i,j < L}(f_c(\mX))_{L(i-1)+j}
\subsection{Image Encoder}
To keep the high spatial resolution of the feature maps and retain the information of the image details, we only use the early three convolution layers of ResNet \cite{2016arXiv160305027H}. 
In this way, we can keep the original resolution of images.
%and a RELU layer as  the image encoder. , which is denoted as $B$. 
Then, we use a projection layer $f_p$ to generate the visual representations for each pixel. 
Third, the image feature map is spatially divided into $m$ sliding windows, with $L\times L$ pixels for each window.    
In this way, the representation of the entire image is as follows, 
\begin{align}
 \omega = [\omega_{1},\omega_{2},...,\omega_{m}] = f_p\big((\mathrm{ResNet}(\sigma)\big),\end{align}
where $\sigma$ is the input image and  $\omega_{k} \in R^{L \times L \times d}$ 
denotes the $k$-th window.

% Each window is represented as $\mI_{wk} \in \R^{L \times L \times d}$, where $d$ is the depth of each patch. 

% Which are scanned by a sliding window and each window is spatially divided into $L \times L$ patches. Each image $\mI$ contains $m$ windows and the representation of the whole image $\mI_w$ can be expressed as follows.
% \begin{align}
%  \mI_w = [\mI_{w1},\mI_{w2},...,\mI_{wm}] = f_p((B(\mI))\end{align}

%Given an image \bm{$\mathrm{I}$}, we use several blocks containing  to preprocess the image, which denoted as $B$. During preprocessing, these blocks maintain the size of the initial image to retain each pixel's information as much as possible. We further apply a RELU function to ensure each value of the pixel to be positive. 

\subsection{Multimodal Sarcasm Fusion}
\label{sec:fusion}
During the process of multimodal fusion, we add the document representation $\varpi$ to each window $\omega_{k}$. The result of addition is denoted as $\hat{\omega}_{k}$,  %$\m\hat{I_{wk}}$
\begin{align}
\hat{\omega}_{k} = \varpi + \omega_{k}\quad k = 0,1,...,m
\end{align}
We apply four stages in Swin-Transformer \cite{liu2021Swin} to deeply fuse the two modalities. 
Specifically, each stage contains one patch merging layer and several blocks containing mechanisms of shifted window attention, which calculates the attention between each element in each shifted window with little computational complexity. By doing so, interactions are built between each word of the document and each image pixel without adding additional calculations. The output of this method delivers multimodal fine-grained features that could be applied to sarcasm understanding.

\section{Experiments} 
% \HH{(start)}
\subsection{Evaluation Tasks}
%Aiming at the challenges proposed by our dataset,

To evaluate our model, we perform two MSU tasks, i.e., sarcasm detection and sarcasm localization. 
%two different multimodal tasks to evaluate our model's performance, and we redefine ``sarcasm object localization'' a little to meet the actual needs.
\newline\textbf{Sarcasm detection}: Sarcasm detection aims to identify whether visual or verbal irony exists in the given sample. This task can be formulated as a binary classification problem. 
%which needs to align the visual and linguistic features to find sarcasm.
\newline\textbf{Sarcasm localization}: Sarcasm localization aims to find out the sarcastic clues or objects in a document with textual spans, as well as in the paired image with bounding boxes. 
% into two categories according to the type of the object to be positioned. This task aims to localize the sarcasm object of the news or the news part that simply makes the news sarcasm. Note that this task cannot be regarded as visual grounding due to sarcasm object in text is often with multiple adjective words, or the sentence which flips the polarity the article in a few words (Examples has been shown in Figure \ref{fig:example}).
\subsection{Implementation Details and Settings}
We employ the pre-trained uncased BERT-base \cite{devlin-etal-2019-bert} as the text encoder. For sarcasm localization, we use a linear layer to predict whether a word token is sarcastic in the text and employ YoloX \cite{2021arXiv210708430G} as the head network to output the bounding box of the sarcastic object or region. 
For sarcasm detection and  textual sarcastic localization, we use the binary cross entropy loss function.
For visual sarcastic localization, we employ the CIoU loss function \cite{9523600}. 
We train our model with a single NVIDIA RTX 3090 GPU. 
The learning rate is set to $0.001$ and $0.01$ for sarcasm detection and localization, respectively.
We employ AdamW \cite{article} as the optimizer. The dataset is randomly split into 70\%, 20\%, and 10\% for training, validation, and testing. 
The previous Swin-Transformer has three settings including $Tiny$, $Small$, and $Base$ \cite{liu2021Swin}. 
For the baseline, we configure Swin-Transformer with the $Tiny$ setting for sarcasm localization, and $Base$ for the detection task, as such two settings perform best among all three settings in corresponding tasks.  More details are available in Appendix: Implementation Details and Settings. We repeat experiments for $5$ times with different random seeds and report both mean and variance values. 

\subsection{Evaluation Matrices}
\begin{figure}[b!]
\vspace{-4mm}
    \flushleft    \includegraphics[scale=0.4725]{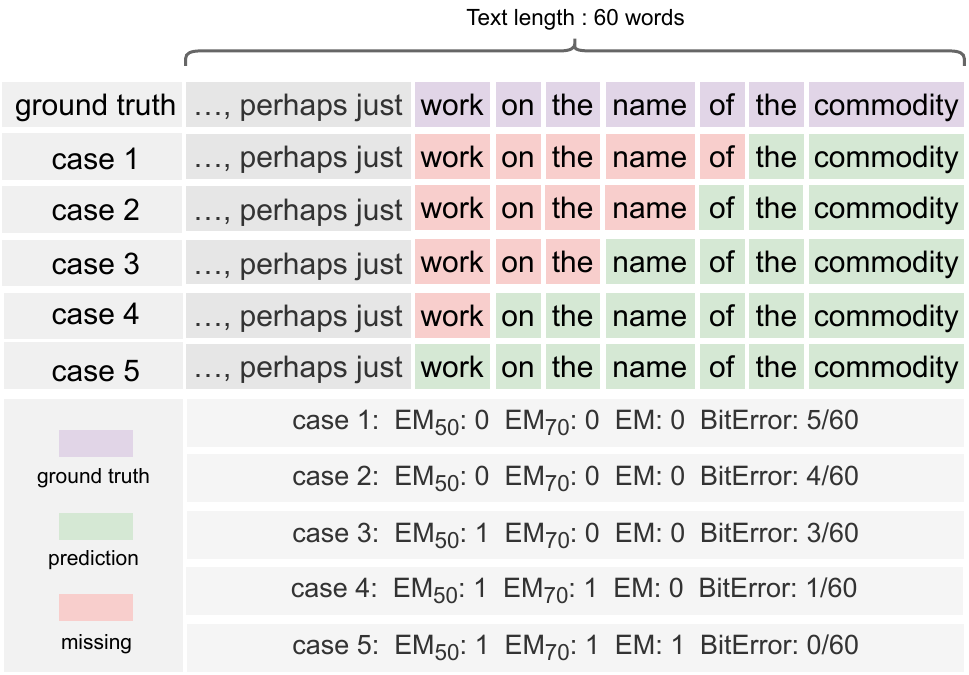}
    \caption{Illustration of the proposed EM$_{50}$, EM$_{70}$, and BitError evaluation metrics for textual sarcastic localization, under the assumption that there are five prediction cases. As case 5 makes the completely correct prediction, its original EM (or EM$_{100}$) can be considered as 1. Although cases $1$ to $4$ yield partially correct predictions, their EM scores are $0$ due to the excessively strict rule of EM.
    Therefore, the original EM cannot properly measure the accuracy. 
    Conversely, our EM$_{50}$ and EM$_{70}$ can better reflect the accuracy from different levels. We additionally introduce a BitError matrix to evaluate predictions based on errors. The proposed evaluation metrics EM$_{50}$, EM$_{70}$ and BitError comprehensively reflect the prediction accuracy of MSU.
    % The example of evaluation matrix. Pre 1 to 5 refer to 5 different models, the words of red background are the targets to predict, the words of purple background are the results of 5 different predictions, the words in red boxes refer to the tokens which are wrongly classified. 
    % EM50, EM70 and EM evaluate the model's performance from 3 different degrees, and BitError is the supplement to the previous three evaluation matrix.
    }
    \label{fig:evaluation matrix}
\end{figure}
\begin{table*}[ht!]

\centering
\scalebox{0.9}{
\resizebox{2\columnwidth}{!}{{
\begin{tabular}{cccccc}
\hline
\multicolumn{6}{c}{\small Sarcasm Localization}                                                        \\ \hline
\multicolumn{1}{c}{}     & \small Model    & \small AP$_{50}$$\uparrow$          & \small F1$_{50}$$\uparrow$                                     &\small  AP$_{60}$$\uparrow$          &\small F1$_{60}$$\uparrow$    \\ \cline{2-6} \hline
                               &\small Swin-Transformer \cite{liu2021Swin}\dag      & \small 21.78\small{($\pm0.09$)}       &\small 21.70\small{($\pm0.10$)}           & \small 6.13\small{($\pm0.21$)}                     &\small 6.10\small{($\pm0.19$)}            \\ 
                               &\small MSTI \cite{wang-etal-2022-multimodal}          &\small 10.21\small{($\pm0.95$)}      &\small 10.17\small{($\pm0.94$)}               &\small 6.31\small{($\pm1.21$)}                     &\small 6.29\small{($\pm1.17$)}            \\ 
                                &\small CLIP \cite{radford2021learning}                 &\small 17.76\small{($\pm0.55$)}      &\small 17.65\small{($\pm0.53$)}                                        &\small 6.23\small{($\pm0.26$)}                     &\small 6.23\small{($\pm0.27$)}            \\ 
\emph{Image} &\small ViLT \cite{pmlr-v139-kim21k}                   &\small 30.73\small{($\pm1.25$)}            &\small 30.68\small{($\pm1.23$)}                             &\small 7.64\small{($\pm0.99$)}                     &\small 7.58\small{($\pm1.47$)}            \\  \cline{2-6} %
                               &\small Ours$_{ST}$            &\small 34.17\small{($\pm2.01$)}              &\small 33.99\small{($\pm2.12$)}                             &\small 10.17\small{($\pm1.21$)}                     &\small 10.05\small{($\pm1.01$)}           \\
                               &\small Ours$_{SS}$            &\small 32.95\small{($\pm1.79$)}             &\small 32.90\small{($\pm1.81$)}                                    &\small 11.68\small{($\pm1.51$)}                     &\small 11.51\small{($\pm2.22$)}           \\
                               &\small Ours$_{SB}$                 &\small \textbf{35.29\small{($\pm2.92$)}}        &\small \textbf{35.24\small{($\pm2.95$)}}                                &\small \textbf{13.74\small{($\pm2.71$)}}                     &\small \textbf{13.67\small{($\pm1.67$)} }          \\ \hline
\multicolumn{1}{c}{}           &\small      Model              &\small EM$_{50}$$\uparrow$                 &\small EM$_{70}$$\uparrow$                                &\small EM$\uparrow$                 &\small BitError$\downarrow $ \\ \hline
                               &\small BERT-base \cite{devlin-etal-2019-bert}\dag     &\small 45.86\small{($\pm0.15$)}                       &\small 36.86\small{($\pm0.26$)}                     &\small 35.77\small{($\pm0.42$)}                     &\small 18.24\small{($\pm0.11$)}           \\  
                                &\small MSTI \cite{wang-etal-2022-multimodal}                 &\small 44.99\small{($\pm0.71$)}                     &\small 35.45\small{($\pm0.73$)}                     &\small 34.10\small{($\pm0.82$)}                     &\small \textbf{14.70\small{($\pm0.20$)}}           \\ 
                                &\small CLIP \cite{radford2021learning}                 &\small 39.34\small{($\pm0.83$)}                     &\small 34.14\small{($\pm0.85$)}                     &\small 33.62\small{($\pm0.99$)}                     &\small 20.47\small{($\pm0.31$)}           \\ 
\emph{Text}   &\small ViLT \cite{pmlr-v139-kim21k}                 &\small 44.09\small{($\pm0.81$)}                     &\small 36.96\small{($\pm0.77$)}                     &\small 36.13\small{($\pm0.75$)}                     &\small 24.64\small{($\pm0.93$)}           \\ \cline{2-6}
                               &\small Ours$_{ST}$             &\small 49.74\small{($\pm0.87$)} &\small 40.17\small{($\pm1.06$)} &\small 38.50\small{($\pm1.04$)} &\small 17.12\small{($\pm0.30$)}          \\
                               &\small Ours$_{SS}$             &\small 50.68\small{($\pm0.92$)} &\small 41.17\small{($\pm1.59$)} &\small \textbf{39.91\small{($\pm0.97$)}} &\small 16.83\small{($\pm0.71$)}           \\
                               &\small Ours$_{SB}$             &\small \textbf{52.19\small{($\pm1.59$)}} &\small \textbf{43.88\small{($\pm0.51$)}} &\small 39.66\small{($\pm0.29$)} &\small 17.67\small{($\pm1.16$)}           \\ \hline
\end{tabular}}}
}
\caption{Comparisons of our method with pre-trained single-modal (denoted as \dag) and multimodal baselines for sarcasm localization. Here we use the subscripts $ST$, $SS$ and $SB$ to represent the different settings of Swin-Transformer in our method, including $Tiny$, $Small$ and $Base$ \cite{liu2021Swin}.}
\vspace{-6mm}
\label{tab:localization}
\end{table*}
For sarcasm detection and sarcasm localization in images, we follow \cite{wang-etal-2022-multimodal} and  \cite{CoCo} to use average precision (AP) and F1 scores for evaluation, respectively, including AP$_{50}$, AP$_{60}$, F1$_{50}$ and F1$_{60}$.  
For textual sarcasm localization, Exact Match (EM) \cite{joshi-etal-2018-sarcasm} is usually employed to measure the prediction accuracy, which is defined as the number of correct predictions  that strictly (100\%) match the boundaries of annotations divided by the total number of predicted samples. 
However, as shown in Figure~\ref{fig:evaluation matrix}, the original EM is too strict to reflect the prediction accuracy. 
Therefore, we introduce three new evaluation matrices as follows, 
% including EM50, EM70 and BitError, which are detailed as follows. %The former two are based on exact   which  We detail the 
%Nevertheless, through the analysis of the tasks and other downstream applications, there is no need to match the boundaries of annotations strictly. EM cannot evaluate the model's performance comprehensively either.
\begin{itemize}
\item \textbf{EM$_{50}$, EM$_{70}$:} We use EM$_{50}$ and EM$_{70}$ to relax the standard EM. They are defined as the number of predictions that match more than $50\%$ and $70\%$ annotations divided by the total number of predicted samples.  The original EM can be seen as EM$_{100}$. 
\item \textbf{BitError:} BitError is the ratio of those wrongly classified tokens to the total number of tokens in a document sample. 
\end{itemize}

\subsection{Sarcasm Detection Results}

% \noindent\textbf{Competitors.} 
\begin{table}[h!]
  \begin{center}
  %\resizebox{1\columnwidth}{!}{
  \scalebox{0.9}{
  \setlength{\tabcolsep}{4pt} %% default is 6pt
    
    \begin{tabular}{cccc} \hline
    %\bottomrule
    \multicolumn{4}{c}{Sarcasm Detection}   \\ \hline
    %\midrule
      \textbf{Model} & \textbf{Acc} & \textbf{Pre} & \textbf{F1-score}\\ \hline
      %\midrule
      BERT-base \cite{devlin-etal-2019-bert}\dag & 87.12 & 77.61 & 86.51 \\ \hline
      Swin-Transformer \cite{liu2021Swin}\dag & 74.83 & 67.57 & 61.51 \\ \hline
      CMGCN \cite{liang-etal-2022-multi}& 88.12 & 78.11 & 75.23 \\ \hline
      CLIP \cite{radford2021learning} & 96.19 & 78.99 & 77.62 \\ \hline
      ViLT \cite{pmlr-v139-kim21k} & 93.15 & 69.03  & 41.44 \\ \hline
      Ours$_{ST}$ & 96.40 & 76.71 & 80.16 \\ 
      Ours$_{SS}$ & 96.82 & 78.10 & 82.75 \\ 
      Ours$_{SB}$ & \textbf{97.83} & \textbf{81.20} & \textbf{87.25} \\ \hline
      %\bottomrule
    \end{tabular}
    }
    \caption{ BERT and Swin-Transformer are based on the single modality (denoted as \dag). Here the subscripts $ST$, $SS$, and $SB$ refer to $Tiny$, $Small$, and $Base$ settings of Swin-Transformer, respectively.
    }
    \label{tab:detection}
    \vspace{-6mm}
  \end{center}
\end{table}

In this paper, we compare our method with BERT-base (text-only)~\cite{devlin-etal-2019-bert}, Swin Transformer (image-only)~\cite{liu2021Swin}, CLIP~\cite{radford2021learning}, Vision-and-Language Transformer (ViLT)~\cite{pmlr-v139-kim21k} and CMGCN~\cite{liu2021Swin}, which detects sarcasm by the object types.
For CLIP and ViLT, we first concatenate the global image and text features and then perform binary classification. 
Our model employs three settings of Swin-Transformer to extract image representations respectively.

As shown in Table~\ref{tab:detection}, our method with Swin-Transformer of $Base$ achieves the best accuracy, demonstrating its superiority in sarcasm detection. 
Because the single-modal BERT-base and Swin-Transformer do not comprehensively exploit the image and text information, they only achieve suboptimal results.
Moreover, the proposed method also outperforms the multimodal CLIP and ViLT models. 
This is because our method is based on more fine-grained visual signals, and the sliding-window-based Transformer mechanism can better capture the sarcasm clues. More detailed comparisons are available in Appendix: Analysis of Swin-Transformer under different settings.

\subsection{Sarcasm Localization Results}
For the sarcasm localization, we additionally include the sentence-level Multimodal Sarcasm Target Identification (MSTI)\cite{wang-etal-2022-multimodal} method, which aims at finding sarcasm clues in tweets. The experimental results are shown in Table\ref{tab:localization}. 
Our method outperforms those existing methods in both visual and textual sarcasm localization in terms of AP-, F1- and EM-based metrics.   
For example, in textual sarcasm localization, our method (SB) surpasses CLIP, ViLT, and MSTI by 12.85\%, 7.20\%, and 8.10\%, respectively, in terms of EM$_{50}$. This shows the effectiveness of the proposed method in localizing nuanced clues in images or long text, and also implies the meaningfulness of collecting such a document-level benchmark. For textual sarcasm localization, we also observe that our method (SS) is the second best and performs slightly lower than MSTI by $2.13$ points in terms of BitError. Nevertheless, our method (SS) significantly outperforms MSTI in terms of EM-based metrics, e.g., $5.81$ and $7.20$ points respectively in terms of EM and EM$_{50}$. The results suggest that our method can achieve a good balance between coverage and precision during the localization. We will further investigate such an interesting finding in the future. We also provide a case study to visually demonstrate how our model performs multimodal sarcasm localization. Due to the space limitation, we give such an illustration in Appendix: Case Study.

%Visual and textual clues of sarcasm are usually relatively small in document and are not evenly distributed in images and texts. Because our model is able to observe small details of images and texts and effectively align and match the visual and textual clues, it is able to achieve promising results. Hence, our method can properly tackle the new challenges of MSU discussed in Section \ref{sec:motivation}. 
%In contrast, although the sentence-level MSTI achieves high accuracy in sarcasm localization on tweets, it fails to show superiority to document-level MSU, implying the meaningfulness collecting such benchmark and data. 

% We run our experimental $5$ times under the same random seed in each round for baselines and our model. We report the mean value and the variance in Table \ref{tab:localization} for visual and textual sarcasm localization. 

\subsection{Attention Visualization}

\begin{figure}[h]
    \centering
    \includegraphics[width=\linewidth]{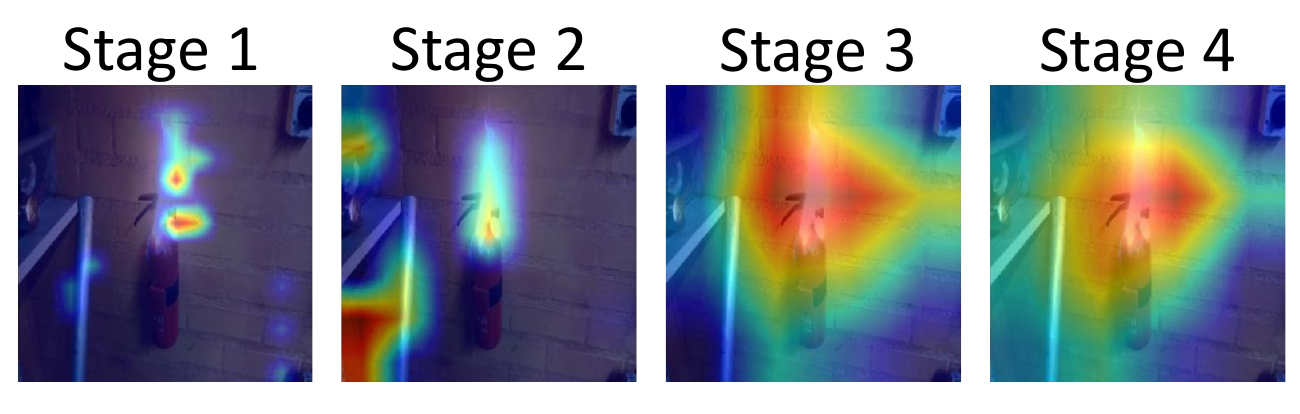}
    \caption{Attention visualization of our method.}
    \label{fig:visualization of attention}
    \vspace{-4mm}
\end{figure}

Figure \ref{fig:visualization of attention} depicts the attention map generated by our method. Our model focuses more on the fire extinguisher after four alignment stages, as discussed in Section \ref{sec:fusion}. This demonstrates the superiority of our approach in capturing fine-grained textual and image clues in document-level multimodal news.

% \begin{figure}[h]
%     \centering
%     \includegraphics[width=76mm]{imgs/example1.pdf}
%     \caption{\textbf{Example of a piece of sarcasm news.} The sarcasm of this news segment is that the fire extinguisher, a tool that should be used to put out fires, has caught fire. This ironic point can raise doubts about the quality and reliability of fire extinguishers.}
%     \label{fig:introexample}
% \end{figure}

% \begin{figure}[h]
% \vspace{-3mm}
%     \flushleft    \includegraphics[scale=0.1725]{imgs/Visual example.pdf}
%     \caption{
%     % The example of evaluation matrix. Pre 1 to 5 refer to 5 different models, the words of red background are the targets to predict, the words of purple background are the results of 5 different predictions, the words in red boxes refer to the tokens which are wrongly classified. 
%     % EM50, EM70 and EM evaluate the model's performance from 3 different degrees, and BitError is the supplement to the previous three evaluation matrix.
%     }
%     \label{fig:attention visualization}
% \end{figure}

\subsection{Ablation Study}

%\noindent\textbf{Impact of Multimodal Modeling}

\subsubsection{Impact of Modalities on MSU.}

%In this paper, we collect a multimodal benchmark for the MSU task. 
This section investigates the influence of visual and textual modalities on MSU. As shown in Table~\ref{tab:abl-det} and Table~\ref{tab:abl-loc}, combining the two modalities significantly improves the detection and localization accuracy. Such comparisons confirm our hypothesis that multimodal cues can benefit sarcasm understanding at the very beginning of Section \ref{sec:intro}.

\begin{table}[]
\small%
\centering
\scalebox{1.05}{
\begin{tabular}{lccc} \hline
%\bottomrule
\multicolumn{4}{c}{Sarcasm Detection}                        \\ \hline
Modality    & Accuracy$\uparrow$     & Precision$\uparrow$     & F1-score$\uparrow$  \\  \hline
Image & 74.83 & 67.57 & 61.51   \\  
Text & 87.12 & 77.61 & 86.51    \\  
Image + Text    & \textbf{97.83} & \textbf{81.20} & \textbf{87.25}   \\  \hline
\end{tabular}
}
\vspace{-2mm}
\caption{Impact of modalities.}
\vspace{-4mm}
\label{tab:abl-det}
\end{table}

\begin{table}[]
\small
\centering
\scalebox{1.02}{
\begin{tabular}{lcccc}\hline %\bottomrule\hline
\multicolumn{5}{c}{Sarcasm Localization}           \\\hline
Modality     & AP$_{50}$$\uparrow$     & F1$_{50}$$\uparrow$   & AP$_{60}$$\uparrow$    & F1$_{60}$$\uparrow$    \\ \hline
Image  & 21.78 & 21.70  & 6.13 & 6.10   \\
Image + Text      &  \textbf{35.29}       &\textbf{35.24}    & \textbf{13.18} & \textbf{13.12}  \\ \hline \hline
 Modality          & EM$_{50}$$\uparrow$                 & EM$_{70}$$\uparrow$                                & EM$\uparrow$                 & BitError$\downarrow $ \\ \hline
Text & 45.86 & 36.86 & 35.77 & 18.24  \\
Text + Image   & \textbf{49.74} & \textbf{40.17} & \textbf{38.50} & \textbf{17.12}  \\ \hline
\end{tabular}}

\vspace{-2mm}
\caption{Impact of modalities.}
\vspace{-4mm}
\label{tab:abl-loc}
\end{table}

\begin{table}[h]\small%

\setlength{\tabcolsep}{6.25pt}
\centering
%\scalebox{0.98}{
\begin{tabular}{l c c c}\hline%\bottomrule
\multicolumn{4}{c}{Sarcasm Detection}                        \\ \hline 
Fusion Method    & Accuracy$\uparrow$     & Precision$\uparrow$     & F1-score$\uparrow$  \\  \hline
Concatenation & 90.27 & 79.69 & 86.62    \\  
The Proposed   & \textbf{97.83} & \textbf{81.20} & \textbf{87.25}   \\  \hline
\end{tabular}%}
\vspace{-2mm}
\caption{Effectiveness of our image-text fusion method.}
\label{tab:abl-fuse-dec}
\vspace{-3mm}
\end{table}

\begin{table}[h!]\small%
\vspace{-1mm}
\centering
%\scalebox{0.98}{
%\resizebox{1\columnwidth}{!}{{
% \renewcommand{\arraystretch}{1.2}
\setlength{\tabcolsep}{6pt}
\begin{tabular}{lcccc}\hline%\bottomrule
\multicolumn{5}{c}{Sarcasm Localization}           \\\hline
Fusion Method   & AP$_{50}$$\uparrow$    & F1$_{50}$$\uparrow$     & AP$_{60}$$\uparrow$   & F1$_{60}$$\uparrow$    \\ \hline
Concatenation  & 21.51 & 21.40  & 5.31 &  5.30   \\
The Proposed   & \textbf{41.04}   & \textbf{40.87}  & \textbf{27.08} & \textbf{27.01}  \\ \hline \hline
Fusion Method         & EM$_{50}$$\uparrow$                 & EM$_{70}$$\uparrow$                                & EM$\uparrow$                 & BitError$\downarrow $ \\ \hline
Concatenation  & 43.16 & 33.33 & 32.37 & 18.36 \\
The Proposed     & \textbf{52.19} & \textbf{42.33} & \textbf{40.59} & \textbf{17.19}  \\ \hline 
\end{tabular}%}}}
\vspace{-2mm}
\caption{Effectiveness of our image-text fusion method.}
\label{tab:abl-fuse-loc}
\vspace{-6mm}
\end{table}
\subsubsection{Impact of Image-Text Fusion Method.}
As discussed in Section \ref{sec:fusion}, this paper presents a new method to fuse visual and textual modalities for MSU. 
To evaluate the effectiveness of the fusion method, we compare ours with a baseline where the encoded image pixels and text tokens are directly concatenated for sarcasm detection and localization. As shown in Table~\ref{tab:abl-fuse-dec} and Table~\ref{tab:abl-fuse-loc}, the proposed fusion method can better capture and align the visual and textual sarcasm clues and achieves better accuracy than the sample concatenation fusion method. These findings show the superiority of our method for the challenging MSU task.

\subsection{DocMSU with Large Language Models}
% We conduct experiments on large language models (LLMs), including GPT-4 \cite{openai2023gpt4}, VideoChat \cite{2023videochat}, Otter \cite{li2023otter} and mPLUG-Owl \cite{ye2023mplugowl}. 
%  While for the challenging ones, LLMs are still unable to understand sarcasm precisely. Detailed results have been shown in Appendix F.
% We access GPT-4 with text inputs and the other LLMs with image-text pairs. We test both MSU tasks with human interactions and show the detailed results in Appendix ??.   

% results when the text contains more complete satirical information, while they cannot answer the satirical object and its cause well when the text does not contain satire and only the combination of both is satirical. We also found that LLMs give good explanations when the satire involves complex cultural knowledge, social context, etc.

We conducted experiments on large language models (LLMs), including GPT-4 \cite{openai2023gpt4}, VideoChat \cite{2023videochat}, Otter \cite{li2023otter}, and mPLUG-Owl \cite{ye2023mplugowl}. 
For the instances with obvious satirical clues, LLMs can yield satisfied performance. While for the challenging ones,
LLMs still struggle to accurately comprehend sarcasm. Detailed results are presented in Appedix: Tests on LLMs. 
Particularly, we observe that LLMs encounter difficulty in accurately identifying the satirical object and its underlying cause when the text does not obviously indicate satire.
Furthermore, it shows that LLMs excel in providing insightful explanations when the news involves intricate cultural knowledge and social context.

\iffalse
\begin{table}[]
\resizebox{1\columnwidth}{!}{
\begin{tabular}{l|lllll}\bottomrule
\multicolumn{6}{c}{Sarcasm Localization}  \\\hline
\multicolumn{1}{c|}{}       & Model      & AP$_{50}$$\uparrow$    & F1$_{50}$$\uparrow$    & AP$_{60}$$\uparrow$    & AP$_{70}$$\uparrow$    \\
\multicolumn{1}{c|}{}       & Ours       & 75.90\% & 75.70\% & 43.02\% & 13.00\%  \\
\multicolumn{1}{c|}{Visual} & w/o blocks & 74.02\% & 73.94\% & 42.08\% & 11.90\%  \\
\multicolumn{1}{c|}{}       & w/o fusion & 70.50\% & 70.40\% & 29.23\% &  9.41\%  \\ \bottomrule
                            & Model      & EM$_{50}$$\uparrow$    & EM$_{70}$$\uparrow$    & EM$\uparrow$          
& BitError$\downarrow$ \\
                            & Ours       & 66.59\% & 60.56\% & 58.46\% & 18.18\%  \\
Text                        & w/o blocks & 61.24\% & 51.92\% & 42.73\% & 20.15\%  \\
                            & w/o fusion & 56.29\% & 35.58\% & 33.57\% & 26.74\%  \\\bottomrule
\end{tabular}
}
\caption{Effectiveness of the proposed image-text fusion method.}
\label{tab:abl-fuse-loc}
\end{table}
\fi

\noindent

\section{Conclusion}
This paper presents DocMSU, a new benchmark for the challenging document-level multimodal sarcasm understanding in the news field. Compared with the existing ones, our DocMSU is more comprehensive, more challenging, and involves higher-quality annotations. We believe our DocMSU will encourage the exploration and development of various downstream tasks for document-level multimodal sarcasm perception closer to real-world applications. The proposed DocMSU also introduces two new challenges. This motivates us to present a new model that aims to capture fine-grained visual sarcastic clues in the image and word-level clues in documents and align them for effective fusion. Experiments on two MSU tasks show the effectiveness of our model on the challenging DocMSU. Future work could focus on MSU across various cultures, as well as the interesting expressive differences between males and females.

\section{Ethical Statement}
We have the copyright of contents collected from three websites, including TheOnion, UNNews, and NewsThump, as these sites automatically grant copyright for users who follow their online rules. 
We carefully study these rules and strictly conform to the requirements during data collection and annotation. 
These online copyright requirements are available on the above websites. 
To further fortify ethical compliance, we will take the following steps:
1). Implementing rigorous data anonymization techniques to safeguard personal information. 
2). Ensuring transparency about the data sources and collection methods in our revised manuscript. 
3). Committing to ongoing scrutiny and readiness to remove or alter data that may be deemed ethically inappropriate or has been
collected from sources that do not provide the necessary authorization. 
4). Developing an online agreement to require every user of the dataset strictly conform to the rules of the
websites from which we collected the data.
\section{Acknowledgments}
% This work was partly supported by the joint funds for Regional Innovation and Development of the National Natural Science Foundation of China (No. U21A20449).
% This work was also partly supported by the Beijing Natural Science Foundation under Grant M21037
% This work was also partly supported by the Fundamental Research Funds for the Central Universities under Grant 2242022k60006.
This work was partially supported by the joint funds for Regional Innovation and Development of the National Natural Science Foundation of China (No. U21A20449), the Beijing Natural Science Foundation under Grant M21037, and the Fundamental Research Funds for the Central Universities under Grant 2242022k60006.
\bibliography{aaai24}

\begin{thebibliography}{35}
\providecommand{\natexlab}[1]{#1}

\bibitem[{Baltrušaitis, Ahuja, and Morency(2019)}]{8269806}
Baltrušaitis, T.; Ahuja, C.; and Morency, L.-P. 2019.
\newblock Multimodal Machine Learning: A Survey and Taxonomy.
\newblock \emph{IEEE Transactions on Pattern Analysis and Machine Intelligence}, 41(2): 423--443.

\bibitem[{Barbieri, Saggion, and Ronzano(2014)}]{barbieri-etal-2014-modelling}
Barbieri, F.; Saggion, H.; and Ronzano, F. 2014.
\newblock Modelling Sarcasm in {T}witter, a Novel Approach.
\newblock 50--58.

\bibitem[{Cai, Cai, and Wan(2019)}]{cai-etal-2019-multi}
Cai, Y.; Cai, H.; and Wan, X. 2019.
\newblock Multi-Modal Sarcasm Detection in {T}witter with Hierarchical Fusion Model.
\newblock 2506--2515.

\bibitem[{Castro et~al.(2019)Castro, Hazarika, P{\'e}rez-Rosas, Zimmermann, Mihalcea, and Poria}]{castro-etal-2019-towards}
Castro, S.; Hazarika, D.; P{\'e}rez-Rosas, V.; Zimmermann, R.; Mihalcea, R.; and Poria, S. 2019.
\newblock Towards Multimodal Sarcasm Detection (An {\_}{O}bviously{\_} Perfect Paper).
\newblock In \emph{Proceedings of the 57th Annual Meeting of the Association for Computational Linguistics}, 4619--4629. Florence, Italy: Association for Computational Linguistics.

\bibitem[{Devlin et~al.(2019)Devlin, Chang, Lee, and Toutanova}]{devlin-etal-2019-bert}
Devlin, J.; Chang, M.-W.; Lee, K.; and Toutanova, K. 2019.
\newblock {BERT}: Pre-training of Deep Bidirectional Transformers for Language Understanding.
\newblock 4171--4186.

\bibitem[{{Ge} et~al.(2021){Ge}, {Liu}, {Wang}, {Li}, and {Sun}}]{2021arXiv210708430G}
{Ge}, Z.; {Liu}, S.; {Wang}, F.; {Li}, Z.; and {Sun}, J. 2021.
\newblock {YOLOX: Exceeding YOLO Series in 2021}.
\newblock \emph{arXiv e-prints}, arXiv:2107.08430.

\bibitem[{{He} et~al.(2016){He}, {Zhang}, {Ren}, and {Sun}}]{2016arXiv160305027H}
{He}, K.; {Zhang}, X.; {Ren}, S.; and {Sun}, J. 2016.
\newblock {Identity Mappings in Deep Residual Networks}.
\newblock \emph{ArXiv preprint}, abs/1603.05027.

\bibitem[{Hu et~al.(2019)Hu, Zhang, Xie, and Lin}]{hu2019local}
Hu, H.; Zhang, Z.; Xie, Z.; and Lin, S. 2019.
\newblock Local Relation Networks for Image Recognition.
\newblock In \emph{Proceedings of the IEEE/CVF International Conference on Computer Vision (ICCV)}, 3464--3473.

\bibitem[{Joshi et~al.(2018)Joshi, Goel, Bhattacharyya, and Carman}]{joshi-etal-2018-sarcasm}
Joshi, A.; Goel, P.; Bhattacharyya, P.; and Carman, M. 2018.
\newblock Sarcasm Target Identification: Dataset and An Introductory Approach.

\bibitem[{Joshi, Sharma, and Bhattacharyya(2015)}]{joshi-etal-2015-harnessing}
Joshi, A.; Sharma, V.; and Bhattacharyya, P. 2015.
\newblock Harnessing Context Incongruity for Sarcasm Detection.
\newblock In \emph{Proceedings of the 53rd Annual Meeting of the Association for Computational Linguistics and the 7th International Joint Conference on Natural Language Processing (Volume 2: Short Papers)}, 757--762. Beijing, China: Association for Computational Linguistics.

\bibitem[{Khodak, Saunshi, and Vodrahalli(2018)}]{khodak-etal-2018-large}
Khodak, M.; Saunshi, N.; and Vodrahalli, K. 2018.
\newblock A Large Self-Annotated Corpus for Sarcasm.
\newblock In \emph{Proceedings of the Eleventh International Conference on Language Resources and Evaluation ({LREC} 2018)}. Miyazaki, Japan: European Language Resources Association (ELRA).

\bibitem[{Kim, Son, and Kim(2021)}]{pmlr-v139-kim21k}
Kim, W.; Son, B.; and Kim, I. 2021.
\newblock ViLT: Vision-and-Language Transformer Without Convolution or Region Supervision.
\newblock In Meila, M.; and Zhang, T., eds., \emph{Proceedings of the 38th International Conference on Machine Learning}, volume 139 of \emph{Proceedings of Machine Learning Research}, 5583--5594. PMLR.

\bibitem[{Kingma and Ba(2014)}]{article}
Kingma, D.; and Ba, J. 2014.
\newblock Adam: A Method for Stochastic Optimization.
\newblock \emph{International Conference on Learning Representations}.

\bibitem[{Li et~al.(2023{\natexlab{a}})Li, Zhang, Chen, Wang, Yang, and Liu}]{li2023otter}
Li, B.; Zhang, Y.; Chen, L.; Wang, J.; Yang, J.; and Liu, Z. 2023{\natexlab{a}}.
\newblock Otter: A Multi-Modal Model with In-Context Instruction Tuning.
\newblock \emph{arXiv preprint arXiv:2305.03726}.

\bibitem[{Li et~al.(2023{\natexlab{b}})Li, He, Wang, Li, Wang, Luo, Wang, Wang, and Qiao}]{2023videochat}
Li, K.; He, Y.; Wang, Y.; Li, Y.; Wang, W.; Luo, P.; Wang, Y.; Wang, L.; and Qiao, Y. 2023{\natexlab{b}}.
\newblock VideoChat: Chat-Centric Video Understanding.
\newblock \emph{arXiv preprint arXiv:2305.06355}.

\bibitem[{Li et~al.(2020)Li, Yin, Li, Zhang, Hu, Zhang, Wang, Hu, Dong, Wei, Choi, and Gao}]{10.1007/978-3-030-58577-8_8}
Li, X.; Yin, X.; Li, C.; Zhang, P.; Hu, X.; Zhang, L.; Wang, L.; Hu, H.; Dong, L.; Wei, F.; Choi, Y.; and Gao, J. 2020.
\newblock Oscar: Object-Semantics Aligned Pre-training for Vision-Language Tasks.
\newblock In Vedaldi, A.; Bischof, H.; Brox, T.; and Frahm, J.-M., eds., \emph{Computer Vision -- ECCV 2020}, 121--137. Cham: Springer International Publishing.
\newblock ISBN 978-3-030-58577-8.

\bibitem[{Liang et~al.(2022)Liang, Lou, Li, Yang, Gui, He, Pei, and Xu}]{liang-etal-2022-multi}
Liang, B.; Lou, C.; Li, X.; Yang, M.; Gui, L.; He, Y.; Pei, W.; and Xu, R. 2022.
\newblock Multi-Modal Sarcasm Detection via Cross-Modal Graph Convolutional Network.
\newblock In \emph{Proceedings of the 60th Annual Meeting of the Association for Computational Linguistics (Volume 1: Long Papers)}, 1767--1777. Dublin, Ireland: Association for Computational Linguistics.

\bibitem[{Lin et~al.(2014)Lin, Maire, Belongie, Hays, Perona, Ramanan, Doll{\'a}r, and Zitnick}]{CoCo}
Lin, T.-Y.; Maire, M.; Belongie, S.; Hays, J.; Perona, P.; Ramanan, D.; Doll{\'a}r, P.; and Zitnick, C.~L. 2014.
\newblock Microsoft COCO: Common Objects in Context.
\newblock In Fleet, D.; Pajdla, T.; Schiele, B.; and Tuytelaars, T., eds., \emph{Computer Vision -- ECCV 2014}, 740--755. Cham: Springer International Publishing.
\newblock ISBN 978-3-319-10602-1.

\bibitem[{Liu et~al.(2021)Liu, Lin, Cao, Hu, Wei, Zhang, Lin, and Guo}]{liu2021Swin}
Liu, Z.; Lin, Y.; Cao, Y.; Hu, H.; Wei, Y.; Zhang, Z.; Lin, S.; and Guo, B. 2021.
\newblock Swin Transformer: Hierarchical Vision Transformer using Shifted Windows.
\newblock In \emph{Proceedings of the IEEE/CVF International Conference on Computer Vision (ICCV)}.

\bibitem[{Mao et~al.(2021)Mao, Shen, Yu, and Cai}]{Mao_Shen_Yu_Cai_2021}
Mao, Y.; Shen, Y.; Yu, C.; and Cai, L. 2021.
\newblock A Joint Training Dual-MRC Framework for Aspect Based Sentiment Analysis.
\newblock \emph{Proceedings of the AAAI Conference on Artificial Intelligence}, 35(15): 13543--13551.

\bibitem[{Nakayama et~al.(2018)Nakayama, Kubo, Kamura, Taniguchi, and Liang}]{doccano}
Nakayama, H.; Kubo, T.; Kamura, J.; Taniguchi, Y.; and Liang, X. 2018.
\newblock {doccano}: Text Annotation Tool for Human.
\newblock Software available from https://github.com/doccano/doccano.

\bibitem[{OpenAI(2023)}]{openai2023gpt4}
OpenAI. 2023.
\newblock GPT-4 Technical Report.
\newblock arXiv:2303.08774.

\bibitem[{Oprea and Magdy(2020)}]{oprea-magdy-2020-isarcasm}
Oprea, S.; and Magdy, W. 2020.
\newblock i{S}arcasm: A Dataset of Intended Sarcasm.
\newblock In \emph{Proceedings of the 58th Annual Meeting of the Association for Computational Linguistics}, 1279--1289. Online: Association for Computational Linguistics.

\bibitem[{Pt{\'a}{\v{c}}ek, Habernal, and Hong(2014)}]{ptacek-etal-2014-sarcasm}
Pt{\'a}{\v{c}}ek, T.; Habernal, I.; and Hong, J. 2014.
\newblock Sarcasm Detection on {C}zech and {E}nglish {T}witter.
\newblock 213--223.

\bibitem[{Qi et~al.(2023)Qi, Bu, Cao, Ji, Shui, Xiao, Wang, and Chua}]{Qi_Bu_Cao_Ji_Shui_Xiao_Wang_Chua_2023}
Qi, P.; Bu, Y.; Cao, J.; Ji, W.; Shui, R.; Xiao, J.; Wang, D.; and Chua, T.-S. 2023.
\newblock FakeSV: A Multimodal Benchmark with Rich Social Context for Fake News Detection on Short Video Platforms.
\newblock \emph{Proceedings of the AAAI Conference on Artificial Intelligence}, 37(12): 14444--14452.

\bibitem[{Radford et~al.(2021)Radford, Kim, Hallacy, Ramesh, Goh, Agarwal, Sastry, Askell, Mishkin, Clark et~al.}]{radford2021learning}
Radford, A.; Kim, J.~W.; Hallacy, C.; Ramesh, A.; Goh, G.; Agarwal, S.; Sastry, G.; Askell, A.; Mishkin, P.; Clark, J.; et~al. 2021.
\newblock Learning transferable visual models from natural language supervision.
\newblock In \emph{International Conference on Machine Learning}, 8748--8763. PMLR.

\bibitem[{Riloff et~al.(2013)Riloff, Qadir, Surve, De~Silva, Gilbert, and Huang}]{riloff-etal-2013-sarcasm}
Riloff, E.; Qadir, A.; Surve, P.; De~Silva, L.; Gilbert, N.; and Huang, R. 2013.
\newblock Sarcasm as Contrast between a Positive Sentiment and Negative Situation.
\newblock 704--714.

\bibitem[{Shu et~al.(2017)Shu, Sliva, Wang, Tang, and Liu}]{KaiShu2017FakeND}
Shu, K.; Sliva, A.; Wang, S.; Tang, J.; and Liu, H. 2017.
\newblock Fake News Detection on Social Media: A Data Mining Perspective.
\newblock \emph{Sigkdd Explorations}.

\bibitem[{Wang et~al.(2022)Wang, Sun, Liu, Shao, and Zheng}]{wang-etal-2022-multimodal}
Wang, J.; Sun, L.; Liu, Y.; Shao, M.; and Zheng, Z. 2022.
\newblock Multimodal Sarcasm Target Identification in Tweets.
\newblock 8164--8175.

\bibitem[{Wilson(2006)}]{WILSON20061722}
Wilson, D. 2006.
\newblock The pragmatics of verbal irony: Echo or pretence?
\newblock \emph{Lingua}, 116(10): 1722--1743.
\newblock Language in Mind: A Tribute to Neil Smith on the Occasion of his Retirement.

\bibitem[{Ye et~al.(2023)Ye, Xu, Xu, Ye, Yan, Zhou, Wang, Hu, Shi, Shi, Jiang, Li, Xu, Chen, Tian, Qi, Zhang, and Huang}]{ye2023mplugowl}
Ye, Q.; Xu, H.; Xu, G.; Ye, J.; Yan, M.; Zhou, Y.; Wang, J.; Hu, A.; Shi, P.; Shi, Y.; Jiang, C.; Li, C.; Xu, Y.; Chen, H.; Tian, J.; Qi, Q.; Zhang, J.; and Huang, F. 2023.
\newblock mPLUG-Owl: Modularization Empowers Large Language Models with Multimodality.
\newblock arXiv:2304.14178.

\bibitem[{Ying et~al.(2022)Ying, Hu, Zhou, Qian, Zeng, and Ge}]{Ying2022BootstrappingMR}
Ying, Q.; Hu, X.; Zhou, Y.; Qian, Z.; Zeng, D.; and Ge, S. 2022.
\newblock Bootstrapping Multi-View Representations for Fake News Detection.
\newblock In \emph{AAAI Conference on Artificial Intelligence}.

\bibitem[{Yu et~al.(2016)Yu, Jiang, Wang, Cao, and Huang}]{Yu2016UnitBoxAA}
Yu, J.; Jiang, Y.; Wang, Z.; Cao, Z.; and Huang, T.~S. 2016.
\newblock UnitBox: An Advanced Object Detection Network.
\newblock \emph{Proceedings of the 24th ACM international conference on Multimedia}.

\bibitem[{Zhang, Zhang, and Fu(2016)}]{zhang-etal-2016-tweet}
Zhang, M.; Zhang, Y.; and Fu, G. 2016.
\newblock Tweet Sarcasm Detection Using Deep Neural Network.
\newblock In \emph{Proceedings of {COLING} 2016, the 26th International Conference on Computational Linguistics: Technical Papers}, 2449--2460. Osaka, Japan: The COLING 2016 Organizing Committee.

\bibitem[{Zheng et~al.(2022)Zheng, Wang, Ren, Liu, Ye, Hu, and Zuo}]{9523600}
Zheng, Z.; Wang, P.; Ren, D.; Liu, W.; Ye, R.; Hu, Q.; and Zuo, W. 2022.
\newblock Enhancing Geometric Factors in Model Learning and Inference for Object Detection and Instance Segmentation.
\newblock \emph{IEEE Transactions on Cybernetics}, 52(8): 8574--8586.

\end{thebibliography}

\end{document}